\documentclass[runningheads]{llncs}
\usepackage[T1]{fontenc}
\usepackage{placeins}
\usepackage{graphicx}
\usepackage{amsmath}
\usepackage{amssymb}
\usepackage{amsfonts}
\usepackage{booktabs}
\usepackage{multirow}
\usepackage{algorithm}
\usepackage{algorithmic}
\usepackage{xcolor}
\usepackage{tikz}
\usetikzlibrary{shapes.geometric, arrows}
\usepackage{enumitem}
\tikzstyle{startstop} = [rectangle, rounded corners, text centered, draw=black]
\tikzstyle{io} = [trapezium, trapezium stretches=true, trapezium left angle=70, trapezium right angle=110, text centered, draw=black, text width=3cm]
\tikzstyle{process} = [rectangle, draw=black, text width=1.5cm, text centered]
\tikzstyle{decision} = [diamond, draw=black, text width=1.5cm, text centered]
\tikzstyle{arrow} = [thick,->,>=stealth]

\providecommand{\keywords}[1]{\textit{Keywords:} #1}
\begin{document}

\title{LogTinyLLM: Tiny Large Language Models Based Contextual Log Anomaly Detection}
\titlerunning{LogTinyLLM}

\author{Isaiah Thompson Ocansey\inst{1} \and Ritwik Bhattacharya\inst{1} \and Tanmay Sen\inst{2}}
%
% \authorrunning{Isaiah Thompson Ocansey et al.}
\institute{Department of Mathematical Sciences, University of Texas at El Paso, El Paso, TX, USA \\
\email{iocansey@miners.utep.edu, ritwik@utep.edu}
\and
SQC \& OR Unit, Indian Statistical Institute, Kolkata, India \\
\email{tanmay.sen@isical.ac.in}}

% \authorrunning{IT et al.}
\maketitle
\thispagestyle{empty} % Ensure first page has no numbering

% Used for displaying a sample figure. If possible, figure files should
% If you use the hyperref package, please uncomment the following two lines
% to display URLs in blue roman font according to Springer's eBook style:
%\usepackage{color}
%\renewcommand\UrlFont{\color{blue}\rmfamily}
%\urlstyle{rm}
%
         % typeset the header of the contribution
%

\begin{abstract}
\small Log anomaly detection using traditional rule based or deep learning based methods is often challenging due to the large volumne and highly complex nature of log sequence.  So effective way of detection of anomalous seqeunce of logs is crucial for system  maintenance and development. This paper proposes parameter efficient finetuning specifically low rank adaptation (LoRA) and adapter based approaches for finding contextual anomalies in sequence of logs in large log data set. It compares different  tiny large language models (LLMs) on the Thunderbird dataset. The results show that LoRA based finetuning provides substantial performance improvements of 18 to 19 percentage over LogBert based full finetuing approach, achieving accuracy scores between 97.76\% and 98.83\% compared to 79.37\%. 
% The finetuned DeepSeek R1-Distill-Qwen-1.5B model excels with an F1-score of 98.56\% and precision-recall characteristics above 94\%. 
%This approach reduces costs associated with full finetuning making it a practical solution for enterprise log monitoring systems.


% \keywords{First keyword  \and Second keyword \and Another keyword.}
\end{abstract}

% \begin{Keywords}

% \end{Keywords}

\section{Introduction}
\small System logs are crucial for monitoring system activities and events, providing insights into performance and security. However, detecting anomalies in these logs are challenging due to their large volume and often unstructured nature. Over the years, traditional machine learning methods, such as Principal Component Analysis (PCA) and one class support vector machines (OC-SVM)  \cite{1b}, have been employed to detect log anomalies. However, these approaches often fail to capture the temporal information within log sequences, limiting their effectiveness. Deep learning techniques, particularly recurrent neural networks (RNNs) \cite{schmidt1912recurrent}, have improved upon these methods by preserving the temporal information of the tokens (log lines) in the log sequences. To enhance RNNs, the DeepLog model \cite{10}  introduced Long Short-Term Memory (LSTM) networks, which effectively model temporal relationships. Despite of being a key benchmark, DeepLog struggles with long-range dependencies. The introduction of the transformer architecture in the 2017 paper "Attention Is All You Need" \cite{11} revolutionized log anomaly detection. Transformer models utilize multi-headed self-attention, allowing for the nuanced interpretation of log sequences while preserving contextual information, thereby improving the accuracy of the anomaly detection. The LogBert model \cite{12}, based on the Transformer architecture, achieved improved results over DeepLog in log anomaly detection.  A decoder based unsupervised approach LogGPT \cite{13} is also employed, which leverages the GPT architecture for sequential log key prediction. The model captures inherent dependencies in log sequences through the next-token prediction and incorporates a reinforcement learning strategy with top-$k$ reward mechanisms to improve detection accuracy. While LogGPT demonstrates superior performance compared to BERT-based approaches, both LogBERT and LogGPT face computational scalability challenges inherent to large transformer models. Traditional full fine tuning of large language models (LLMs) for log anomaly detection can be expensive in terms of computational resources and time.  Advancements in parameter efficient fine tuning (PEFT) techniques, such as Low Rank Adaptation (LoRA)\cite{hu2022lora}, have revolutionized adapting large pre-trained models for specific anomaly detection tasks. LoRA strategically reduces the number of trainable parameters by freezing the original weights of the model and training only a small subset of additional parameters. This methodology significantly lowers memory consumption and computational cost while preserving the model's high performance, making it especially suitable for anomaly detection in large scale system logs \cite{3}. In this article we propose LogTinyLLM, a parameter-efficient fine-tuning (PEFT) methods with tiny large language models, utilizing low rank adaptation (LoRA) and adapter methods to reduce training cost and model size. This framework aims for a lightweight, resource efficient solution for log anomaly detection. We present our findings in the subsequent sections.

\section{Methodology}
\subsection{Log parsing}
Log parsing transforms raw log data into a structured format, making it easier to identify anomalies using the drain algorithm \cite{he2017drain}. This algorithm extracts "log keys" (or message types) from the raw logs through a tree-like structure with a fixed depth, organizing them systematically. After the logs are parsed, the log keys are organized into sequences using sliding window technique. This approach enables the analysis of system activity over time. The extracted log sequences are utilized with tiny large language models to capture the inherent semantics present in the log data.
\subsection{ Tiny Large Language Models (Tiny LLMs) }
Fine tuning pretraind LLMs requires substantial computational resources, which can be very costly. To reduce training expenses, we utilize tiny LLMs optimized to provide similar performance while significantly lowering the training cost. This approach makes it more affordable for organizations with limited resources and easy deployment into applications. The following are the tiny LLMs used in this research: \\

% \subsubsection{Opt-1.3b}
\noindent \textbf{OPT-1.3B: } Zhang et al.~\cite{zhang2022opt} introduced open pre-trained transformers (OPT), a family of auto regressive causal language models (decoder only) developed by Meta AI, which range from 125 million to 175 billion parameters which is open source alternative to GPT-3 for language modeling, text generation, and other NLP tasks. The model supports a context window that is appropriate for modeling sequential dependencies in log data and provides reliable inference performance for log classification. It consists of 24 layers, each containing 32 attention heads, an embedding size of 2048 and vocabulary size of 50272. 
% It is a mid sized model within this family, demonstrates competitive performance compared to models of similar scale. The open access nature of OPT-1.3B allows for flexible integration and tuning, facilitating domain specific adaptation without the high costs associated with training a model from scratch. T

\noindent \textbf{Phi-1.5:} Li et al.~\cite{li2023textbooks} introduced \textit{Phi-1.5}, a decoder only transformer based language model with 1.3 billion parameters developed by Microsoft and designed for efficient reasoning, coding, and math problem-solving. It was trained on a high quality dataset consisting of 30 billion tokens, which includes 7 billion tokens from Phi-1's filtered code corpus and 20 billion synthetically generated tokens modeled after textbooks. The model has 24 transformer layers, each with 32 attention heads and a hidden size of 2048, using Rotary Positional Embeddings (RoPE) with a rotary dimension of 32. 
% It uses FlashAttention and a code friendly tokenizer, making it ideal for real time, low resource applications without the need for instruction tuning or RLHF. Despite its small size, Phi-1.5 performs competitively with much larger models like LLaMA-2 7B, while being more lightweight and resource efficient. 

\noindent \textbf{TinyLlama-1.1B:} Zhang et al.~\cite{zhang2024tinyllama} introduced \textit{TinyLlama}, a open source tiny large language model with 1.1 billion parameters based on the LLaMA 2 architecture. It built as a decoder only transformer with 22 transformer layers, 32 attention heads, and a hidden size of 2048. The latest version, TinyLlama v1.1, was trained on up to 2 trillion tokens, utilizing a multi-stage pre-training process that incorporated domain-specific corpora, including SlimPajama, StarCoder, ProofPile, and Skypile. 

% This paper utilizes TinyLlama-1.1B due to its excellent performance to size ratio and efficient architecture. With a context window of 2,048 tokens and optimized features such as Rotary Positional Embeddings, Grouped-Query Attention, SwiGLU activations, and FlashAttention-2, it provides  fast and accurate detection of anomalies in  log sequences. 

\noindent \textbf{DeepSeek-R1-Distill-Qwen-1.5B:} Guo et al.~\cite{guo2025deepsee} introduced \textit{DeepSeek-R1}, a family of first generation reasoning optimized language models developed using reinforcement learning (RL) without supervised fine tuning (SFT) as a preliminary step. DeepSeek-R1-Distill-Qwen-1.5B model is distilled from the larger DeepSeek-R1 model, with the base architecture of Qwen-2.5-Math-1.5B. It consists of 28 transformer layers, each with a hidden size of 2048, and uses Grouped-Query Attention (GQA), where 32 attention heads are split into 12 query heads and 2 shared key value heads, reducing memory cost while preserving performance. The model employs RoPE to encode position information within queries and keys, supporting a maximum context length of 32,768 tokens. It uses SwiGLU activation in the feedforward blocks for better non-linearity. 
% The feedforward network has an intermediate size of 5632. 
% This paper utilizes \textit{DeepSeek-R1-Distill-Qwen-1.5B} due to its efficient architecture and proven performance in reasoning tasks. The model benefits from distilled reasoning behaviors inherited from larger base models, allowing it to process structured log sequences effectively. 

\section{Parameter Efficient Fine Tuning (PEFT)}
PEFT builds on the concept of finetuning by introducing additional layers or parameters to the pre-trained model, focusing training exclusively on the newly added components and keeping most of the original model’s parameters frozen. 
% PEFT addresses the high computational costs associated with traditional finetuning by limiting the trainable parameters to these augmented layers. This selective updating significantly reduces the memory and processing resources required.  PEFT achieves this efficiency without compromising the model’s performance. 
In this paper, we explore low-rank adaptation (LoRa) and adapter based PEFT methods.

\begin{figure}[hbt!]
  \centering
  \includegraphics[height=6cm,width=\textwidth
    % width=0.6\columnwidth,  % reduced width
    % clip,
    % trim=5 5 5 5             % optional: trims whitespace
  ]{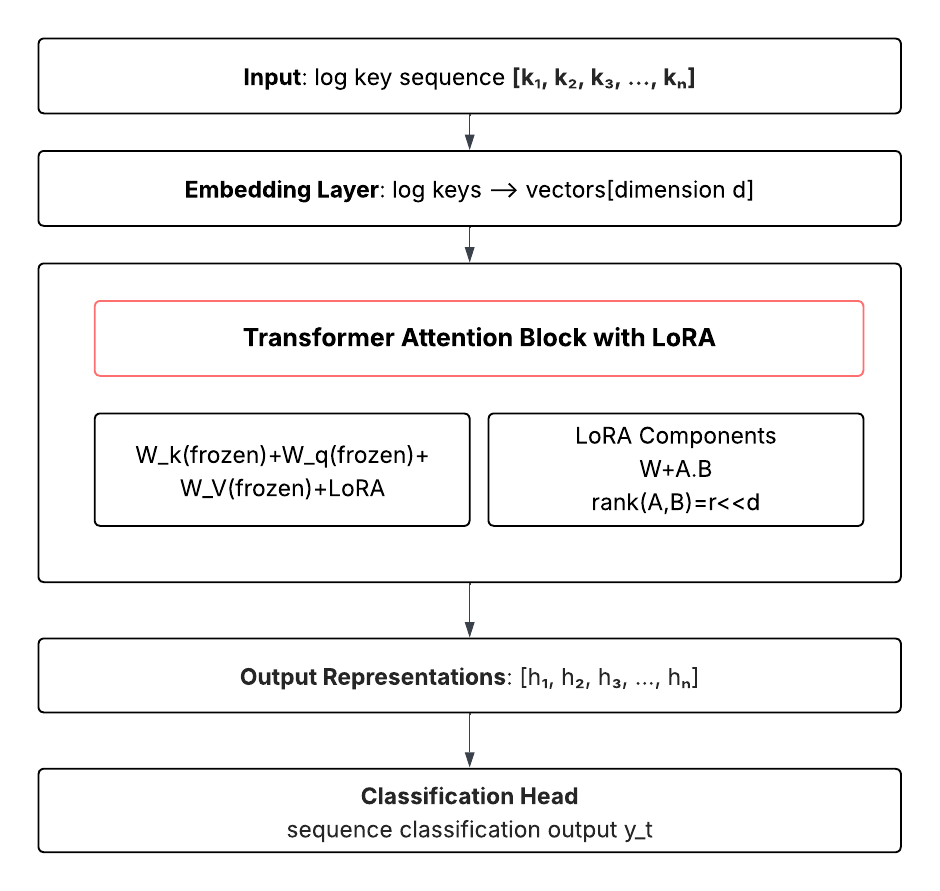} 
  \caption{LoRA architecture for LogTinyLLMs.}
  \label{fig:lora_architecture}
\end{figure}

\subsection{Low-Rank Adaptation (LoRA)}
We consider Low Rank Adaptation (LoRA) \cite{hu2022lora}, a peft based  method  for finetuning pretrained large language models for log sequence anomaly detection task. Specifically, instead of updating the full weight matrices in the attention mechanism, we keep the original pretrained weights fixed and update it with small trainable low rank matrices into the query, key, and value projections. These low rank matrices adapt the new task without changing most of the original parameters. Let the input sequence be represented as $H \in \mathbb{R}^{T \times d}$, where each row $h_t = e_t + p_t $ combines the log token embedding $ e_t \in \mathbb{R}^d $ and its positional encoding $ p_t \in \mathbb{R}^d$. For each transformer layer and attention head $l$, the standard self-attention operation is given by  \cite{1b}, $\text{Attention}(Q, K, V) = \text{softmax} \left( \frac{Q K^\top}{\sqrt{d_k}} \right)V$, with the query, key, and value matrices computed as:
$Q = H W_Q^l, \quad K = H W_K^l, \quad V = H W_V^l$, 
where $W_Q^l, W_K^l, W_V^l \in \mathbb{R}^{d \times d_k}$ are the learnable projection matrices in the pretrained model. Instead of directly updating these large projection matrices, LoRA introduces a low rank decomposition of the update as $W^l_{\text{adapted}} = W^l + \Delta W^l = W^l + \alpha \cdot B^l A^l$, where  $B^l = \mathbf{0}$  and $A^l \sim \mathcal{N}(0, \sigma^2)$. 
% where:
% \begin{itemize}
%     \item \( A^l \in \mathbb{R}^{r \times d_k} \), \( B^l \in \mathbb{R}^{d \times r} \) are trainable low-rank matrices,
%     \item \( r \ll \min(d, d_k) \) is the LoRA rank, and
%     \item \( \alpha \) is a scaling factor to balance the magnitude of the adaptation.
% \end{itemize}
We apply this decomposition to each of the projection matrices:
$Q = H (W_Q^l + \alpha \cdot B_Q A_Q),~
K = H (W_K^l + \alpha \cdot B_K A_K),~
V = H (W_V^l + \alpha \cdot B_V A_V).$ The proposed methodology is given in Algorithm \ref{alg:Lora-sequence-classification} and the flowchart is given in Fig \ref{fig:lora_architecture}.
\begin{algorithm}[hbt!]
\caption{LoRA Fine Tuning  for Log Sequence Anomaly Detection}
\label{alg:Lora-sequence-classification}
\begin{algorithmic}[1]
\STATE \textbf{Input:}
\footnotesize% \begin{itemize}
    Pretrained model \( M_{\text{pretrained}} \), 
    Training data \( D = \{(k_i, a_i, y_i)\}_{i=1}^N \), $k_i$ is a sequence of log keys, N is the number of log sequence, 
    LoRA rank \( r \), scaling factor $\alpha$, learning rate \( \eta \),  and number of epochs \( T \)
    Set of target layers \( \mathcal{L}_{\text{target}} \subseteq \{1, \dots, L\} \)
% \end{itemize}

\STATE \textbf{Output:} Fine-tuned model \( M_{\text{LoRA}} \)

\STATE \textcolor{blue}{// Step 1: Freeze all original weights}
\STATE Freeze all parameters of \( M_{\text{pretrained}} \)

\STATE \textcolor{blue}{// Step 2: Inject LoRA into selected attention projection matrices}
\FOR{each layer \( \ell \in \mathcal{L}_{\text{target}} \)}
    \FOR{each projection matrix \( W^\ast_\ell \in \{W^Q_\ell, W^K_\ell, W^V_\ell\} \)}
        \STATE Initialize trainable low-rank matrices \( A^\ast_\ell \in \mathbb{R}^{r \times d},\ B^\ast_\ell \in \mathbb{R}^{d \times r} \)
        \STATE Modify projection: \( W^\ast_\ell \gets W^\ast_\ell + \alpha (B^\ast_\ell A^\ast_\ell) \)
    \ENDFOR
\ENDFOR

\STATE \textcolor{blue}{// Step 3: Add a lightweight classifier head (if needed)}
\STATE Add classifier \( f_{\text{cls}}: \mathbb{R}^d \rightarrow \mathbb{R}^2 \)

\STATE \textcolor{blue}{// Step 4: Optimize only LoRA parameters and classifier}
\FOR{epoch \( t = 1 \) to \( T \)}
    \FOR{each mini-batch \( \{(k_i, a_i, y_i)\} \subset D \)}
        \STATE Compute transformer output: \( \mathbf{H}_i = M_{\text{LoRA}}(k_i, a_i) \)
        \STATE Pooling: \( \mathbf{s}_i = \text{MeanPool}(\mathbf{H}_i, a_i) \)
        \STATE Predict: \( \text{logits}_i = f_{\text{cls}}(\mathbf{s}_i) \)
        % \STATE Compute loss: \( \mathcal{L}_i = \text{CrossEntropy}(y_i, \text{logits}_i) \)
        % \STATE \textcolor{blue}{// Step 5: Probability with softmax}
    \STATE \( P(y_i = c \mid k_i) = \frac{\exp(\text{logits}_{i,c})}{\sum_{j=0}^{1} \exp(\text{logits}_{i,j})} \)

    % \STATE \textcolor{blue}{// Step 6: Weighted binary cross-entropy loss over the batch}
\STATE \(\mathcal{L}_{\text{WCE}} = -\frac{1}{N} \sum_{i=1}^N \sum_{c=0}^{1} w_c y_{i,c} \log(P(y_i = c \mid k_i))\)
        \STATE Backpropagate and update: \( A^\ast_\ell, B^\ast_\ell \), and classifier
    \ENDFOR
\ENDFOR

\STATE \textcolor{blue}{// Step 5: Return fine-tuned model}
\STATE Return \( M_{\text{LoRA}} \)

\end{algorithmic}
\end{algorithm}
\vspace{-0.5cm}
\subsection{Adapter Based Finetuning}
Another parameter efficient fine tuning (PEFT) method, adapter modules, was proposed by \cite{houlsby2019parameter}. In this article, we adopt a simplified version of this approach by adding two trainable linear layers on top of the final hidden layer of a frozen decoder only pretrained model to adapt it for the long sequence of log anomaly detection task. Since only these two additional layers are trained, this method requires fine tuning only a small fraction of the total model parameters, making it computationally efficient while preserving the pretrained knowledge. The proposed methodologies is given in  Algorithm \ref{alg:adapter-sequence-classification}.
\begin{algorithm}[hbt!]
\caption{Adapter based parameter efficient fine tuning for log sequence anomaly detection }
\label{alg:adapter-sequence-classification}
\begin{algorithmic}[1]
\footnotesize \REQUIRE A set of \( N \) log sequences \( \{k_1, \dots, k_N\} \), where \( k_i \in \mathbb{Z}^{\text{seq\_len}} \); attention masks \( \{a_1, \dots, a_N\} \), where \( a_i \in \{0,1\}^{\text{seq\_len}} \); binary labels \( \{y_{1,c}, \dots, y_{N,c}\} \), with \( y_{i,c} \in \{0, 1\} \)

%\ENSURE Predicted logits and total loss

\FOR{each input sequence \( (k_i, a_i) \) in the batch}

    \STATE \textcolor{blue}{// Step 1: Contextual encoding via frozen TinyLLM}
    \STATE \( \mathbf{H}_i = \texttt{TinyLLM}(k_i, a_i) \in \mathbb{R}^{\text{seq\_len} \times d} \)
    \COMMENT{Token-level hidden states from final frozen layer}

    \STATE \textcolor{blue}{// Step 2: Mean pooling for sequence representation}
    \STATE \( \mathbf{s}_i = \frac{1}{\sum_{t=1}^{\text{seq\_len}} a_i^t} \sum_{t=1}^{\text{seq\_len}} a_i^t \cdot \mathbf{H}_i[t] \in \mathbb{R}^{d} \)

    \STATE \textcolor{blue}{// Step 3: Adapter MLP (stacked trainable layers)}
    \STATE \( \mathbf{z}_i^{(1)} = \text{ReLU}(\mathbf{W}_1 \mathbf{s}_i + \mathbf{b}_1), \quad  \mathbf{W}_1 \in \mathbb{R}^{d_{\text{hidden}} \times d_{\text{hidden}}} \)
    \STATE \( \mathbf{z}_i^{(2)} = \text{ReLU}(\mathbf{W}_2 \mathbf{z}_i^{(1)} + \mathbf{b}_2), \quad  \mathbf{W}_2 \in \mathbb{R}^{d_{\text{hidden}} \times d_{\text{hidden}}} \)

    \STATE \textcolor{blue}{// Step 4: Final classification layer}
    \STATE \( \text{logits}_i = \mathbf{W}_{\text{cls}} \mathbf{z}_i^{(2)} + \mathbf{b}_{\text{cls}}, \quad  \mathbf{W}_{\text{cls}} \in \mathbb{R}^{2 \times d_{\text{hidden}}} \)

    \STATE \textcolor{blue}{// Step 5: Probability with softmax}
    \STATE \( P(y_i = c \mid k_i) = \frac{\exp(\text{logits}_{i,c})}{\sum_{j=0}^{1} \exp(\text{logits}_{i,j})} \)

\ENDFOR

\STATE \textcolor{blue}{// Step 6: Binary cross-entropy loss over the batch}
\STATE \(\mathcal{L}_{\text{WCE}} = -\frac{1}{N} \sum_{i=1}^N \sum_{c=0}^{1} w_c y_{i,c} \log(P(y_i = c \mid k_i))\)
\RETURN \( \text{logits}_1, \dots, \text{logits}_N; \mathcal{L}_{\text{WCE}} \)
\end{algorithmic}
\end{algorithm}
% \begin{figure}[H]
%   \centering
%   \includegraphics[%
%     width=\columnwidth,   % fills the entire column
%     clip,
%     trim=5 5 5 5          % tweak to remove any extra whitespace
%   ]{Ito LoRA diagram.pdf} 
%   \caption{LoRA architecture for tiny large language models (Tiny-LLMs).}
%   \label{fig:lora_architecture}
% \end{figure}
\section{Experiment}
\subsection{Dataset}
This research utilizes an open sourced log dataset, Thunderbird \cite{oliner2007supercomputers,zhu2023loghub}, which is a comprehensive collection of logs sourced from the Thunderbird supercomputer located at Sandia National Labs (SNL) in Albuquerque and available on GitHub. This dataset comprises 211,212,192 log entries, with a total raw size of 29.60 GB. In the dataset, log entries that begin with a "-" indicate normal logs, whereas those that do not begin with a "-" in the first column are classified as abnormal logs. 
\begin{figure}[hbt!]
  \centering
  \includegraphics[height=6cm,width=0.8\textwidth]{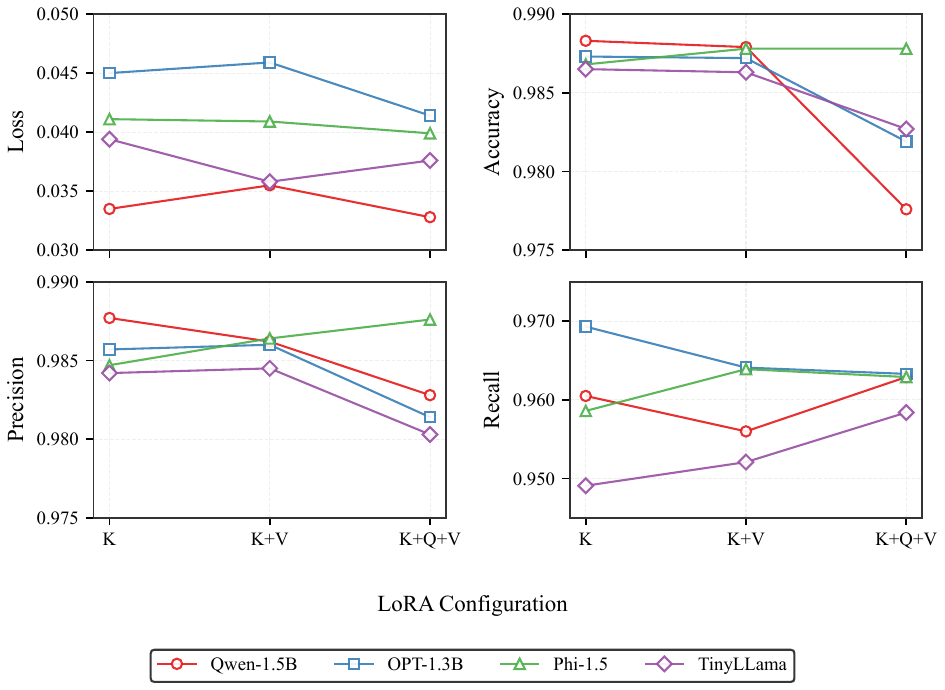}
  \caption{Performance analysis of LoRA on Thunderbird dataset.}
  \label{fig:performance_analysis}
\end{figure}

% \subsection{Evaluation Metrics} 
% The performance of the models are assessed using widely utilized standard classification metrics tailored for binary anomaly detection tasks such as accuracy, recall, precision, and F1 score. All the following expressions will be presented using  $TP$, $TN$, $FP$, and $FN$, which denote true positives, true negatives, false positives, and false negatives, respectively are defined below:
% \noindent\\
% \(\text{Accuracy} = \frac{TP + TN}{TP + TN + FP + FN}, \text{Precision} = \frac{TP}{TP + FP}, 
% \text{Recall} = \frac{TP}{TP + FN}, \\
% \text{$F_1$-Score} = 2 \cdot \frac{\text{Precision} \times \text{Recall}}{\text{Precision} + \text{Recall}}.\)

\begin{table*}[hbt!]
\centering
\caption{Comparative performance analysis of LoRA-based fine-tuning across multiple Tiny-LLMs   on the Thunderbird dataset. }
\label{tab:lora-performance-analysis}
\resizebox{\textwidth}{!}{%
\begin{tabular}{@{}llccccc@{}}
\toprule
\textbf{Model Architecture} & \textbf{LoRA Target Modules} & \textbf{Loss} & \textbf{Accuracy} & \textbf{Precision} & \textbf{Recall} & \textbf{F1-Score} \\
\midrule
\multicolumn{7}{l}{\textit{Single Module Configuration}} \\
\midrule
Microsoft Phi-1.5 & $k_{\text{proj}}$ & 0.0411 & 98.68\% & 98.47\% & 95.86\% & 98.42\% \\
Facebook OPT-1.3B & $k_{\text{proj}}$ & 0.0450 & 98.73\% & 98.57\% & \textbf{96.93\%} & 98.48\% \\
DeepSeek-R1-Distill-Qwen-1.5B & $k_{\text{proj}}$ & \textbf{0.0335} & \textbf{98.83\%} & \textbf{98.77\%} & 96.05\% & \textbf{98.56\%} \\
TinyLLama-1.1B & $k_{\text{proj}}$ & 0.0394 & 98.65\% & 98.42\% & 94.91\% & 98.35\% \\

\midrule
\multicolumn{7}{l}{\textit{Dual Module Configuration}} \\
\midrule
Microsoft Phi-1.5 & $k_{\text{proj}}, v_{\text{proj}}$ & 0.0409 & \textbf{98.78\%} & \textbf{98.64\%} & \textbf{96.39\%} & \textbf{98.51\%} \\
Facebook OPT-1.3B & $k_{\text{proj}}, v_{\text{proj}}$ & 0.0459 & 98.72\% & 98.60\% & 96.41\% & 98.42\% \\
DeepSeek-R1-Distill-Qwen-1.5B & $k_{\text{proj}}, v_{\text{proj}}$ & \textbf{0.0355} & 98.79\% & 98.62\% & 95.60\% & 98.57\% \\
TinyLLama-1.1B & $k_{\text{proj}}, v_{\text{proj}}$ & 0.0358 & 98.63\% & 98.45\% & 95.21\% & 98.34\% \\
\midrule
\multicolumn{7}{l}{\textit{Triple Module Configuration}} \\
\midrule
Microsoft Phi-1.5 & $k_{\text{proj}}, q_{\text{proj}}, v_{\text{proj}}$ & 0.0399 & \textbf{98.78\%} & \textbf{98.76\%} & 96.29\% & \textbf{98.44\%} \\
Facebook OPT-1.3B & $k_{\text{proj}}, q_{\text{proj}}, v_{\text{proj}}$ & 0.0414 & 98.19\% & 98.14\% & \textbf{96.33\%} & 98.16\% \\
DeepSeek-R1-Distill-Qwen-1.5B & $k_{\text{proj}}, q_{\text{proj}}, v_{\text{proj}}$ & \textbf{0.0328} & 97.76\% & 98.28\% & 96.29\% & 97.98\% \\
TinyLLama-1.1B & $k_{\text{proj}}, q_{\text{proj}}, v_{\text{proj}}$ & 0.0376 & 98.27\% & 98.03\% & 95.84\% & 98.12\% \\
\midrule
\multicolumn{7}{l}{\textit{Baseline Comparison}} \\
\midrule
LogBERT\textsuperscript{†} & Full fine-tuning & 0.0958 & 79.37\% & 92.03\% & 51.47\% & 66.02\% \\
\bottomrule
\end{tabular}%
}
\scriptsize All LoRA configurations use rank $r=2$, scaling factor $\alpha=16$, dropout $p=0.05$,  Batch size=2, learning rate=5×10\textsuperscript{-5}, AdamW optimizer and 3 epochs.
% \begin{itemize}
% \footnotesize
% \item[] \textbf{Experimental Setup}: All LoRA configurations use rank $r=2$, scaling factor $\alpha=16$, dropout $p=0.05$.
% \item[] \textbf{Training Configuration}: Batch size=2, learning rate=5×10\textsuperscript{-5}, AdamW optimizer, 3 epochs with early stopping.
% % \end{tablenotes}
% \end{itemize}
\end{table*}

\begin{table*}[hbt!]
\centering
\caption{Comparative Performance Analysis of Adapter Based finetuning on Thunderbird Dataset}
\label{tab:thunderbird-performance}
\resizebox{\textwidth}{!}{%
\begin{tabular}{@{}llcccccc@{}}
\toprule
\multirow{2}{*}{\textbf{Model Family}} & \multirow{2}{*}{\textbf{Architecture}} & \multirow{2}{*}{\textbf{Parameters}} & \multicolumn{5}{c}{\textbf{Performance Metrics}} \\
\cmidrule(lr){4-8}
& & & \textbf{Loss} & \textbf{Accuracy} & \textbf{Precision} & \textbf{Recall} & \textbf{F1-Score} \\
\midrule
TinyLLama & TinyLLama-1.1B-Chat & 1.1B & 0.8420 & 84.17\% & 84.20\% & 82.44\% & 0.8243 \\
Microsoft & Phi-1.5-Base & 1.3B & 0.3716 & 84.30\% & 84.07\% & 84.20\% & 0.8250 \\
Meta AI & OPT-1.3B-Base & 1.3B & 0.2598 & 85.04\% & 86.62\% & 85.04\% & 0.8313 \\
DeepSeek & R1-Distill-Qwen-1.5B & 1.5B & 0.3020 & \textbf{87.95\%} & \textbf{88.79\%} & \textbf{87.95\%} & \textbf{0.8669} \\
LogBERT & BERT-Base & 110M & \textbf{0.0958} & 79.37\% & \textbf{92.03\%} & 51.47\% & 0.6602 \\
\bottomrule
\end{tabular}%
}
% \begin{tablenotes}
% \footnotesize
% \end{tablenotes}
\end{table*}

\subsection{Experimental Results and Discussion}
The results presented in Fig \ref{fig:performance_analysis} clearly demonstrate the effectiveness of the LoRA approach for finetuning various query, key and value projection parameters. The experimental result in Table \ref{tab:lora-performance-analysis}  based on the LoRA method significantly outperforms LogBert for log anomaly detection. It achieves accuracy scores ranging from 97.76\% to 98.83\% and F1 scores between 97.98\% and 98.57\%. In contrast, LogBert only achieves an accuracy of 79.37\% and an F1-score of 66.02\%. The optimal configuration employs single-module adaptation that targets the key projection layer, with the DeepSeek-R1-Distill-Qwen-1.5B model achieving the highest F1-score of 98.56\%. Other models, such as Microsoft Phi-1.5, Facebook OPT-1.3B, and TinyLLama-1.1B, also perform well; however, DeepSeek-R1-Distill-Qwen-1.5B demonstrates the lowest validation loss. LoRA configurations exhibit precision exceeding 98\% and recall rates between 94.91\% and 96.93\%, providing better stability and generalization compared to LogBert. Among the adapter-based models (see Table \ref{tab:thunderbird-performance}), the DeepSeek R1-Distill-Qwen-1.5B model achieved the highest metrics, including an accuracy of 87.95\%, precision of 88.79\%, recall of 87.95\%, and an F1-score of 0.8669, making it the top performer. The other adapter-based models also demonstrated strong performance, with accuracies ranging from 84.17\% to 85.04\%, precision ranging from 84.07\% to 86.62\%, recall from 82.44\% to 85.04\%, and F1-scores ranging from 0.8243 to 0.8313. In contrast, LogBERT, despite having the lowest loss of 0.0958, did not perform as well in other metrics, achieving an accuracy of 79.37\%, precision of 92.03\%, recall of 51.47\%, and an F1-score of 0.6602. These results indicate that adapter based models are more effective than LogBERT for log anomaly detection on the Thunderbird dataset, with the DeepSeek R1-Distill-Qwen-1.5B model achieving the best overall performance.

\subsection{Conclusion}
This paper evaluates two, parameter efficient fine (PEFT) tuning approaches for log sequence anomaly detection, and shows that LoRA works better than LogBert based full fine tuning approach. 
% The proposed methodology achieves state-of-the-art performance on the Thunderbird dataset, with F1 Scores reaching 98.56\%. 
Our study reveals that applying LoRA to just a single layer offers the best trade-off between performance and efficiency. Among the models evaluated, DeepSeek R1-Distill-Qwen-1.5B consistently achieves the highest accuracy on the Thunderbird dataset. 
Adapter-based fine-tuning methods also perform well, though they show a drop of about 10\% in F1 score compared to LoRA. However, they require fewer trainable parameters as well, making them more lightweight than LoRA-based approaches. Finally, our approach avoids full model fine tuning and still achieves high accuracy with lower computational cost, making it suitable for real world use cases with limited resources.

% \FloatBarrier

% include

% \FloatBarrier

% \FloatBarrier
% \begin{figure}
%     \includegraphics[width=\textwidth]{Isaiah3thunderbird_performance_analysis.pdf}
%     \label{fig:performance_analysis}
% \end{figure}

% \begin{figure}[H]
%     \centering
%     \includegraphics[width=\columnwidth]{Isaiah4_thunderbird_f1_comparison.pdf}
    
%     \label{fig:f1_comparison}
% \end{figure}

% \begin{figure}[H]
%     \centering
%     \includegraphics[width=\columnwidth]{Isaiah5_model_comparison.pdf}
%     \label{fig:model_comparison}
% \end{figure}

% \section{Discussion}
% \subsection{LoRA}

 % log anomaly detection
% Furthermore, we find that parameter-efficient fine-tuning methods such as LoRA effectively eliminate the traditional trade-off between precision and recall in anomaly detection. 

%
%



\end{document}